\documentclass[letterpaper, 10 pt, conference]{ieeeconf}  

\IEEEoverridecommandlockouts                              

\usepackage{amsmath} 
\usepackage{amssymb}  
\usepackage{cite}
\usepackage{graphicx}
\usepackage{xcolor}
\usepackage{colortbl}
\usepackage{caption}

\usepackage{multirow} 
\usepackage{pifont}
\newcommand{\cmark}{\ding{51}} 
\usepackage{array}
\newcommand{\Tstrut}{\rule{0pt}{2.6ex}} 
\newcommand{\Bstrut}{\rule[-0.9ex]{0pt}{0pt}} 
\usepackage{xcolor}
\usepackage{colortbl}
\usepackage{caption}
\usepackage{hyperref}

\title{\LARGE \bf
Overlap-Aware Feature Learning for \\ Robust Unsupervised Domain Adaptation for 3D Semantic Segmentation
}

\author{
Junjie Chen\textsuperscript{1},
Yuecong Xu\textsuperscript{2},
Haosheng Li\textsuperscript{1},
Kemi Ding\textsuperscript{1}
\thanks{J. Chen, H. Li and K. Ding are with the School of Automation and Intelligent Manufacturing (AIM), Southern University of Science and Technology, Shenzhen, China. Email: {\tt\small\{12332651, 12332662\} @mail.sustech.edu.cn, dingkm@sustech.edu.cn.}}
\thanks{Y. Xu is with the Department of Electrical and Computer Engineering, National University of Singapore. Email: 
{\tt\small yc.xu@nus.edu.sg}}}

\begin{document}

\hbadness=2000000000
\vbadness=2000000000
\hfuzz=100pt

\maketitle
\thispagestyle{empty}
\pagestyle{empty}

\begin{abstract}
3D point cloud semantic segmentation (PCSS) is a cornerstone for environmental perception in robotic systems and autonomous driving, enabling precise scene understanding through point-wise classification. While unsupervised domain adaptation (UDA) mitigates label scarcity in PCSS, existing methods critically overlook the inherent vulnerability to real-world perturbations (e.g., snow, fog, rain) and adversarial distortions. This work first identifies two intrinsic limitations that undermine current PCSS-UDA robustness: (a) unsupervised features overlap from unaligned boundaries in shared-class regions and (b) feature structure erosion caused by domain-invariant learning that suppresses target-specific patterns. To address the proposed problems, we propose a tripartite framework consisting of: 1) a robustness evaluation model quantifying resilience against adversarial attack/corruption types through robustness metrics; 2) an invertible attention alignment module (IAAM) enabling bidirectional domain mapping while preserving discriminative structure via attention-guided overlap suppression; and 3) a contrastive memory bank with quality-aware contrastive learning that progressively refines pseudo-labels with feature quality for more discriminative representations. Extensive experiments on SynLiDAR-to-SemanticPOSS adaptation demonstrate a maximum mIoU improvement of 14.3\% under adversarial attack.
\end{abstract}


\section{Introduction}
3D semantic segmentation~\cite{milioto2019rangenet++,choy20194d,bian2022unsupervised} is a fundamental task in robotic vision for environmental perception, which involves classifying point-wise labels and is widely used in autonomous driving and robotics. Manual annotation of 3D point clouds requires domain expertise and significant human effort. In light of this, recent works have focused on unsupervised domain adaptation (UDA)~\cite{langer2020domain,gebrehiwot2023t} methods, which aim to transfer knowledge from the labeled source domain to the unlabeled target domain. This improves model generalization across different domains and reduces label dependency. The typical UDA on 3D point cloud semantic segmentation (PCSS-UDA) task assumes corruption-free target domain data. However, this assumption is unrealistic in real-world applications. For example, PCSS-UDA models can dramatically degrade under adversarial weather conditions such as snow, fog, rain~\cite{xiao20233d,eskandar2022unsupervised} and unnoticeable adversarial perturbations~\cite{yang2021exploring,xu2023adversarial}. These issues further widen the domain gap, leading to more intense negative transfer. Yet, the robustness of PCSS-UDA models under adversarial attack and common corruption is currently unexplored.

\begin{figure}[t]
    \centering
    \includegraphics[width=0.9\linewidth]{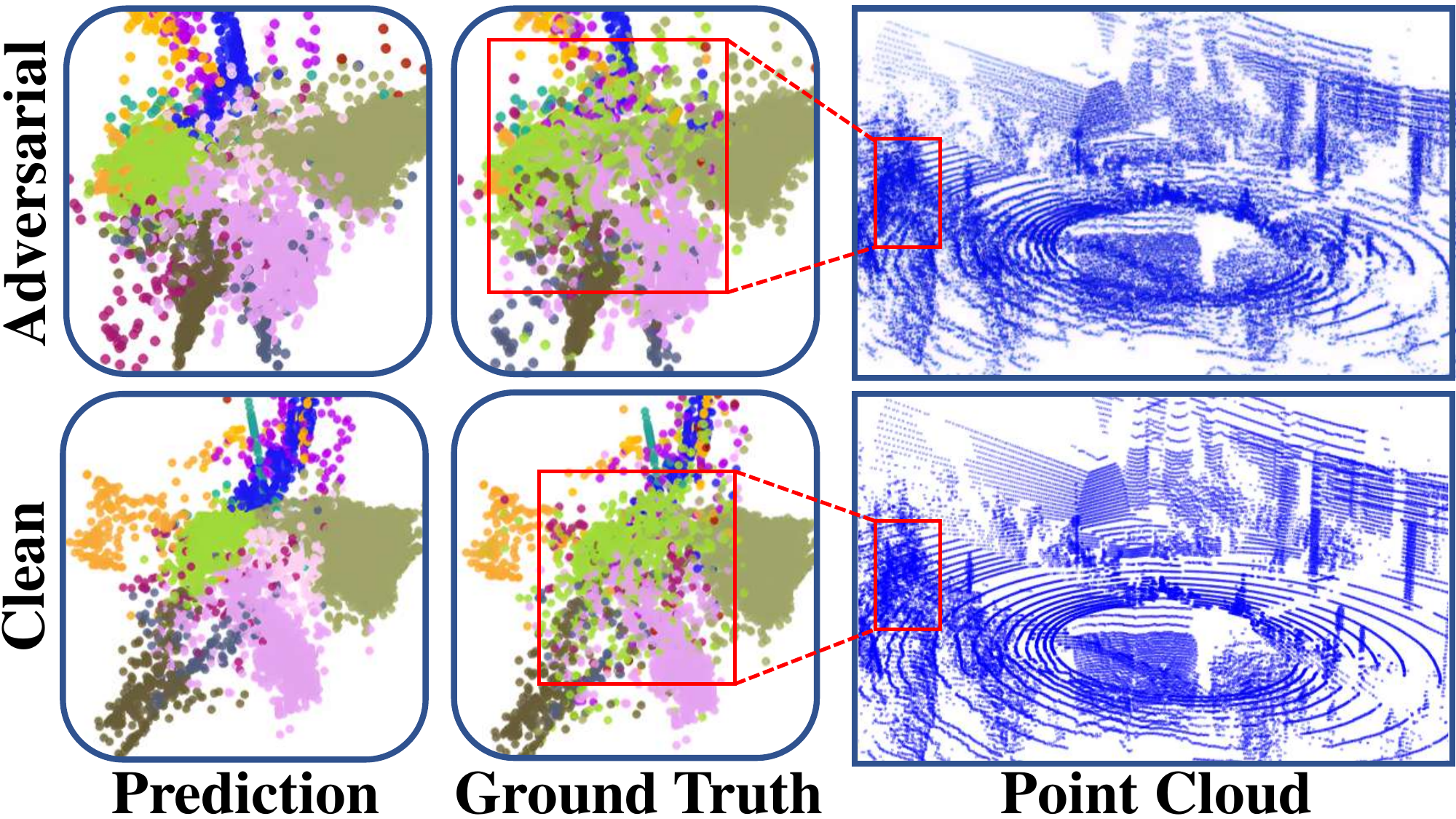}
    \caption{Illustration of adversarial perturbations on feature space representations with Isomap \cite{pai2019dimal}, a widely-used nonlinear dimensionality reduction technique. In the visualization above, we compare the clean (bottom row) and adversarial (top row) feature distributions in the ground truth (right) and predicted (left) spaces. Adversarial perturbations cause a wider dispersion in feature space, where points from different classes increasingly overlap, making classification even more challenging.}
    \label{Fig: Isomap}
    \vspace{-3mm}
\end{figure}

Recent studies in image UDA have indicated that adversarial training methods with adversarial target samples can improve domain generalization and adversarial robustness~\cite{li2021divergence,lo2022exploring,awais2021adversarial}. These methods encourage the model to identify perturbation-invariant features during training. However, adversarial training methods compromise model performance in clean data and lead to longer training time. 
Further, we observe a significant issue in 3D semantic segmentation arises when we treat target domain data as transferable adversarial examples. As shown in Fig. \ref{Fig: Isomap}, adversarial perturbations can cause feature overlap for different classes. Here, the ground truth feature distributions overlap with predicted feature distributions for different classes after adversarial perturbations. Such an overlap causes the model to be transferred to make incorrect predictions with high confidence for unseen target features, thereby obstructing the PCSS-UDA training and degrading model robustness.

We analyzed and identified two reasons for the inferior generalization performance of the PCSS-UDA model in the above scenarios. First, the model has difficulty adapting to overlap classes due to insufficient supervision in the target domain. This significantly hinders the accurate alignment of boundaries between shared classes. Additionally, domain-invariant feature learning further complicates this issue by inadvertently suppressing discriminative feature structure in the target domain. To address the problems mentioned, we devise an Invertible Neural Network (INN)-based Attention Alignment Model (IAAM) that enables bidirectional feature alignment while preserving discriminative feature structure. This is coupled with an overlap suppression loss that dynamically mitigates interference across cross-domain class overlaps. Additionally, we introduce a Quality-guided Contrastive Memory Bank (QC-MB) that employs quality-aware scoring to iteratively distill high-quality and discriminative representations for more reliable pseudo-label generation.

In summary, the contributions of our work are threefold:
\begin{itemize}
    \item We propose a robustness evaluation framework for PCSS-UDA, which benchmarks the model's robustness under adversarial attacks and challenging weather conditions.
    \item To reduce feature overlap and enhance the robustness of the model, we introduce the IAAM and QC-MB modules. These components work in tandem with PCSS-UDA and help to align and exploit more discriminative target features.
    \item Extensive experiments are conducted to validate the effectiveness of our proposed methods, which demonstrate a maximum mIoU improvement of 14.3\% under adversarial attacks for $\text{SynLiDAR} \rightarrow \text{SemanticPOSS}$ adaptation. Our results underline the potential of our approach in mitigating the challenges of adversarial perturbations and domain generalization.
\end{itemize}

\section{Related Work}

\subsection{Unsupervised 3D Semantic Segmentation}
UDA has been extensively explored to transfer knowledge from a labeled source domain to an unlabeled target domain during training. To address the domain shift between the source and target datasets in unsupervised settings, existing work mainly focused on data augmentation methods~\cite{langer2020domain,bevsic2022unsupervised,kim2024density} and feature alignment methods~\cite{kang2019contrastive,zhou2023homeomorphism,yi2021complete,saltori2023compositional}. Data augmentation methods address cross-sensor UDA by transforming high-resolution LiDAR scans into lower-resolution counterparts and mimicking target domain characteristics to facilitate adaptation. Feature alignment methods aim to learn the discriminative distribution alignment of the source and target domain in the latent space. In recent years, self-training methods with pseudo-labeling have gained attention due to their effectiveness in learning transferable representations. Pseudo-labeling methods~\cite{gebrehiwot2023t, saltori2023compositional} iteratively refine predictions in the target domain by leveraging high-confidence outputs from the model, while noisy pseudo-labels can be inevitable. HMA~\cite{zhou2023homeomorphism} is a pioneer work that attempts to apply flow-based INN for conducting domain alignment in two separate spaces, thereby preserving the topological structure of data in both domains, and has shown promise in 2D classification tasks.
These strategies aim to learn more domain-invariant and discriminative features, thus enhancing domain generalizations during training.

\subsection{Adversarial Conditions and Robust UDA}
Previous studies indicate that deep learning models face significant challenges from adversarial attacks. PCSS models are vulnerable to imperceptible perturbations and natural corruptions, which can adversely impact segmentation and domain generalization performance~\cite{liu2022imperceptible,gao2023towards,liu2019extending}. The works in~\cite{xu2023adversarial} and~\cite{liu2022imperceptible} highlight how adversarial perturbations, especially in safety-critical applications such as autonomous driving, can degrade the performance of 3D point cloud models. In addition, adverse weather conditions, such as rain or snow in the target domain~\cite{xiao20233d}, further complicate the domain adaptation process. Existing work for robust UDA has focused on mitigating these challenges and improving model robustness. Li et al.~\cite{li2021divergence} attempt to tackle robust UDA from the perspective of transferable adversarial attacks for 2D classification using adversarial training with target data. In addition, the AR regularization method in~\cite{gao2023towards} proposes minimizing the distribution distance between natural and adversarial data, offering a promising direction to deal with noisy or perturbed target domain data. However, adversarial training is computationally expensive with adversarial sample generation during training. In contrast, existing work on adversarial feature desensitization (AFD)~\cite{bashivan2021adversarial} and robust feature adaptation (RFA)~\cite{awais2021adversarial} propose distilling features that are invariant to adversarial perturbations.  These methods indicate that distilling discriminative features is crucial to enhancing model robustness during UDA. Although these approaches offer a shorter training time than adversarial training, they do not incorporate inherent feature structure within UDA and erode discriminative target features. In contrast to prior work, our work proposes disentangling domain alignment and suppressing overlap across domains for more discriminative learning in PCSS-UDA.


\section{Methodology}
\subsection{Preliminaries and Overview}
In the context of unsupervised domain adaptation on 3D point cloud semantic segmentation (PCSS-UDA) under adversarial conditions of the target domain, the labeled source domain is denoted as \( \mathcal{S} = \{(\mathbf{x}_{i},\mathbf{y}_{i})\}_{i=1}^{|\mathcal{S}|} \) and the unlabeled target domain is denoted as $\mathcal{T} = \{(\mathbf{x}_{i})\}_{i=1}^{|\mathcal{T}|}$. To obtain a point-wise classification of inputs, the pre-trained model is used to map the point cloud $\mathbf{x}_{\mathcal{S}} \in \mathbb{R}^{N \times 3}$ to the inference distributions of classes $p_{\theta}(c|\mathbf{x}_{\mathcal{S}}) \in \mathbb{R}^{N \times C}$, which can be further divided into the feature embedding module $F_{\theta_f}(\mathbf{x}) \in \mathbb{R}^{N \times dim}$ and the classification module $C_{\theta_c}(c | F_{\theta_f}(\mathbf{x})) \in \mathbb{R}^{dim \times C}$. The target model parameters learned by UDA are denoted as $\theta_\mathcal{T}$. For robustness evaluation, the target point cloud may suffer from subtle perturbations: $\hat{\mathbf{x}}_{\mathcal{T}} = \mathbf{x}_{\mathcal{T}} + \delta, \quad \hat{\mathbf{x}}_{\mathcal{T}} \in \mathbb{R}^{N \times 3}$, where $\delta$ denotes the applied perturbation. Our approach aims to achieve robust PCSS-UDA under adversarial conditions, including adversarial attacks and challenging weather conditions. We follow the recent pseudo-labeling technique, CosMix~\cite{saltori2023compositional}, which utilizes a teacher-student framework. The teacher model generates high-confidence pseudo-labels for unlabeled target data and gradually distills knowledge from a pre-trained source model. Our work consists of three key components: a robustness evaluation model, an Invertible Attention Alignment Module with overlap suppression loss (IAAM) and a Quality-guided Contrastive Memory Bank (QC-MB). 

\begin{figure}[t]
    \centering
    \includegraphics[width=0.95\linewidth]{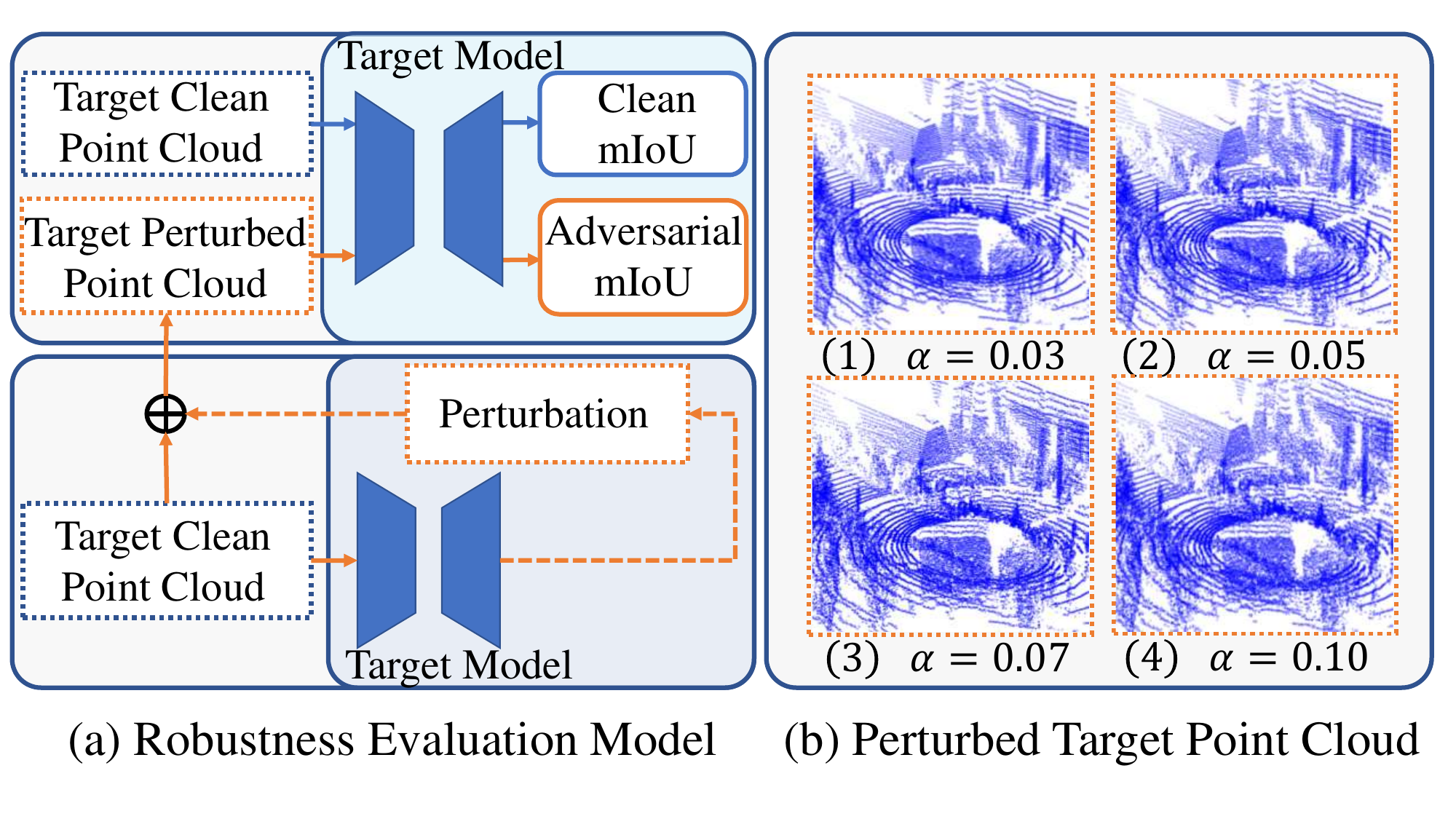}
    \caption{Illustration of our Robustness Evaluation Model, the perturbation is generated based on adversarial objective functions. Figures in (b) display target point clouds with different perturbation intensities.}
    \label{Fig: Evaluation_Robustness_Model}
    \vspace{-3mm}
\end{figure}

\subsection{Robustness Evaluation Model}
\noindent $\textbf{Adversarial attacks.}$ Adversarial attacks typically perturb the original inputs with subtle perturbations but significantly degrade the deep learning model with high confidence~\cite{zhu2021adversarial,rossolini2023real}, which can be divided into white-box attacks and black-box attacks. White-box attacks utilize the parameters of the trained model and data labels to generate adversarial perturbations. In contrast, black-box attacks rely on the transferability of adversarial examples while limited for different model architectures and training datasets~\cite{cheng2019improving,feng2022boosting,chen2024theory}. To accurately assess the worst-case performance degradation under input perturbations, we incorporate the white-box attack with the ground truth target labels and the PCSS-UDA model to generate target adversarial examples while target labels are unavailable during training. For point cloud input $\mathbf{x}$, and its labels $\mathbf{y}$, the adversarial objective is to generate adversarial examples $\hat{\mathbf{x}}$ subject to $\mathbf{M(\hat{\mathbf{x}}) \neq \mathbf{y}}$. In this work, we adopt the projected gradient descent (PGD) attack~\cite{xu2023adversarial} and the iterative fast gradient sign descent (I-FGSM) attack~\cite{liu2019extending} as the adversarial objective function. The generation of I-FGSM adversarial examples is denoted as:
\begin{equation}
\mathbf{x}_{adv}^{t+1} = \mathbf{x}_{adv}^{t} + \alpha \cdot \text{sign}(\nabla_{\mathbf{x}_{adv}^{t}} \mathcal{L}(p_{\theta}(c|\mathbf{x}_{adv}^t), \mathbf{y})),
\end{equation}
in contrast, PGD extends I-FGSM by introducing an additional projection step, ensuring that adversarial perturbations remain within a bounded $l_p$ norm constraint. This prevents perturbations from growing arbitrarily large over iterations. The update rule for PGD attack is denoted as:
\begin{equation}
    \begin{aligned}
       \mathbf{x}_{adv}^{t+1} &= \Pi_{\mathbf{x}+\epsilon} \left( \mathbf{x}_{adv}^{t} + \frac{\alpha \cdot \nabla_{\mathbf{x}_{adv}^{t}} \mathcal{L}(p_{\theta}(c|\mathbf{x}_{adv}^t), \mathbf{y}_)}{\|\nabla_{\mathbf{x}_{adv}^{t}} \mathcal{L}(p_{\theta}(c|\mathbf{x}_{adv}^t), \mathbf{y})\|_2} \right),
    \end{aligned}
\end{equation} 
in which $\mathcal{L}$ is the adversarial loss function, typically using the cross-entropy loss for classification tasks.

\noindent $\textbf{Robustness metric.}$ As illustrated in Fig.~\ref{Fig: Evaluation_Robustness_Model}, different perturbation intensities cause varying degradation of model performance. To quantitatively evaluate the robustness of the PCSS-UDA model, the robustness score is denoted as:
\begin{equation}
\text{Robustness Drop} = \frac{\text{mIoU}_{\text{clean}} - \text{mIoU}_{\text{adv}}}{\text{mIoU}_{\text{clean}}},
\end{equation}
where the Mean Intersection over Union (mIoU) is a widely used evaluation metric in semantic segmentation, which is computed as:
\begin{equation}
    \text{mIoU} = \frac{1}{C} \sum_{i=1}^{C} \frac{|P_i \cap G_i|}{|P_i \cup G_i|},
\end{equation}
where $C$ is the number of classes, $P_i$ and $G_i$ represent the predicted and ground truth sets for class $i$, respectively. The robustness drop metric quantifies the relative performance degradation under adversarial perturbations by comparing mIoU on clean and adversarial examples.

\begin{figure*}[t]
    \centering
    \includegraphics[width=0.8\linewidth]{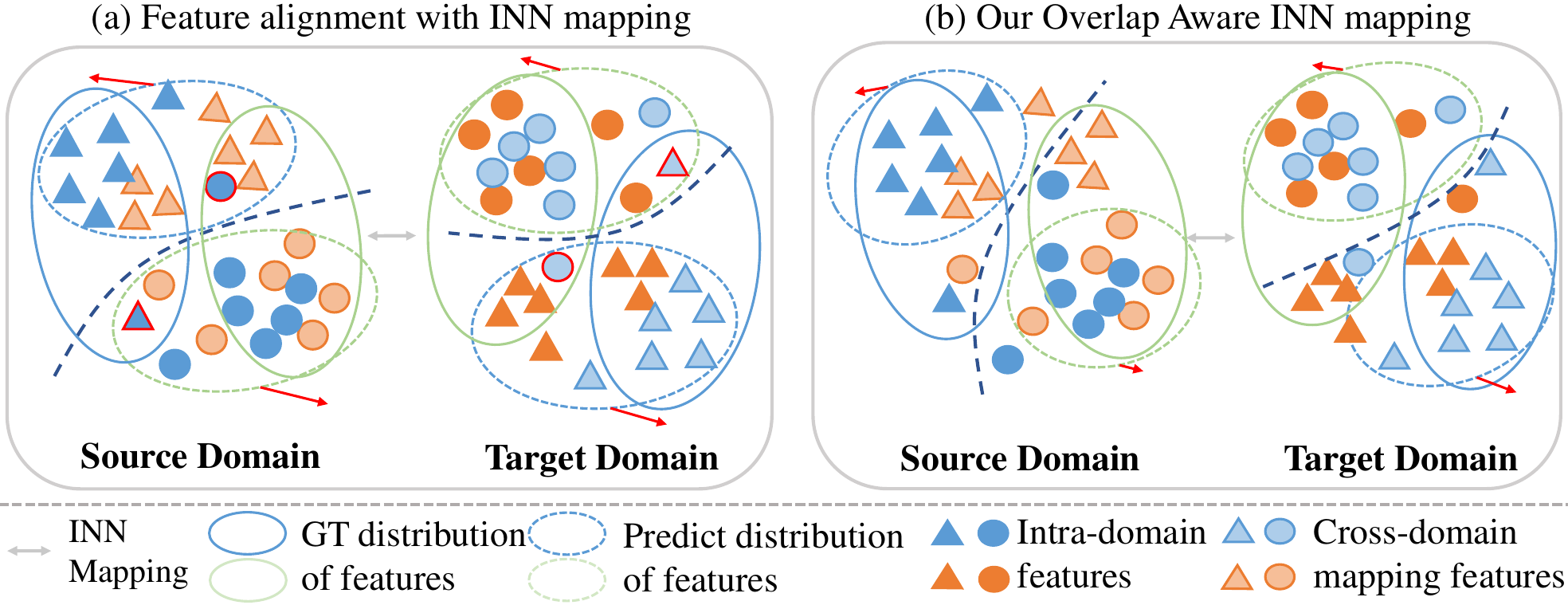}
    \caption{The illustration of our overlap-aware INN mapping method. This approach encourages alignment of the structural target domain features $f_\mathcal{T}$ with the mapping features from the source domain $f_{\mathcal{T} \rightarrow \mathcal{S}}$ and vice versa. The proposed Overlap Loss penalizes the overlap between the predicted feature distributions and the ground truth distributions across domains, thereby minimizing erroneous overlap during the PCSS-UDA task.}
    \label{Fig: OverlapLoss}
    \vspace{-3mm}
\end{figure*}

\subsection{Invertible Attention Alignment Module (IAAM)}

The pseudo-labeling PCSS-UDA method shown in Table.~\ref{table:4-1 lidar2poss under attack} demonstrates low robustness against adversarial perturbations. To further investigate the impact of discriminative target feature structure and feature overlap on model robustness, we propose the IAAM framework, which integrates three synergistic components: (1) a flow-based INN mapping for bidirectional feature alignment; (2) an attention module to generate pseudo-labels for mapped features; and (3) an overlap-aware loss for class overlap suppression. The integrated framework jointly optimizes feature-level mapping and semantic refinement across domains through multitask learning.

\noindent $\textbf{Flow-based bidirectional alignment.}$ To effectively leverage the intrinsic feature structure of the target domain, we employ a flow-based bidirectional alignment mechanism. The INN model learns a bidirectional feature mapping between the source $\mathcal{S}$ and the target $\mathcal{T}$ domains using INN with forward ($f_{\mathcal{S} \rightarrow \mathcal{T}}$) and inverse ($f_{\mathcal{T} \rightarrow \mathcal{S}}$) transformations. This process not only preserves the structural integrity of the target domain, but also enhances feature representations by integrating mapped features with the original representations. The mapping is optimized as follows:
\begin{equation}
\begin{aligned}
\mathcal{L}^m = &\ \mathcal{L}_{\text{mapping}}(f_{\mathcal{S} \rightarrow \mathcal{T}}, f_{\mathcal{T}}) 
+ \mathcal{L}_{\text{mapping}}(f_{\mathcal{T} \rightarrow \mathcal{S}}, f_{\mathcal{S}}) \\
= &\ \frac{1}{2} \left[ \mathcal{W}(f_{\mathcal{S} \rightarrow \mathcal{T}}, f_{\mathcal{T}}) 
+ \mathcal{W}(f_{\mathcal{T} \rightarrow \mathcal{S}}, f_{\mathcal{S}}) \right],
\end{aligned}
\end{equation}
where $\mathcal{L}_\text{mapping}$ denotes the mapping loss of features. Specifically, we utilize $\mathcal{W}(\cdot,\cdot)$, which denotes the optimal transport-based Wasserstein distance~\cite{shen2018wasserstein,nietert2022statistical,nguyen2023energy} computed through Sinkhorn iterations~\cite{cuturi2013sinkhorn}. To preserve semantic integrity during cyclic transformations, we introduce the cyclic consistency loss:
\begin{equation}
\begin{aligned}
\mathcal{L}^c = &\ \frac{1}{2} \|f_{\mathcal{S}} - f_{\mathcal{S} \rightarrow \mathcal{T} \rightarrow \mathcal{S}}\|_p 
+ \frac{1}{2} \|f_{\mathcal{T}} - f_{\mathcal{T} \rightarrow \mathcal{S} \rightarrow \mathcal{T}}\|_p,
\end{aligned}
\end{equation}
where $\|\cdot\|_p$ represents the $\ell_p$-norm (empirically set to $\ell_2$ in our experiments).

\noindent \textbf{Attention-guided pseudo-label generation.} 
Inspired by earlier HMA approaches~\cite{zhou2023homeomorphism} in image classification, we utilize multi-head attention module $\mathcal{A}(\cdot)$ to generate domain-invariant pseudo-labels and class distributions for mapping features through:
\begin{equation}
 \begin{aligned}
     p_{\theta_\mathcal{A}} (c | \cdot) = \text{softmax}\left(\frac{QK^T}{\sqrt{dim}}\right)V,
 \end{aligned}
\end{equation}
where $Q,K,V$ are query, key, and value matrices derived from domain-transformed features. The attention training combines:
\begin{equation}
\begin{aligned}
\mathcal{L}^{\text{att}} 
&= \mathcal{L}_{CE}(\mathcal{A}(f_{\mathcal{S}}), y_{\mathcal{S}}^{\text{gt}})
+ \mathcal{L}_{CE}(\mathcal{A}(f_{\mathcal{S} \rightarrow \mathcal{T}}), y_{\mathcal{S}}^{\text{gt}})\\
& +\|\mathcal{A}(f_{\mathcal{T} \rightarrow \mathcal{S}}), \mathcal{C}_{\theta_c}(c | f_{\mathcal{T}})\|_2,
\end{aligned}
\end{equation}
where $\mathcal{L}_{\text{CE}}$ denotes the cross-entropy loss, and $\mathcal{C}_{\theta_c}(c|\cdot)$ is the classifier output of student model for target domain features. The attention module ensures distribution consistency across original and transformed features. The complete optimization for INN training integrates the mentioned bidirectional alignment and attention generation losses:
\begin{equation}
\mathcal{L}_{\text{INN}} = \mathcal{L}^m + \mathcal{L}^c + \mathcal{L}^{\text{att}},
\end{equation}
this joint optimization enables simultaneous feature alignment and feature-aware pseudo-label generation.

\noindent \textbf{Overlap suppression loss.} To resolve feature overlap for different classes, we propose the overlap loss:
\begin{equation}
\begin{aligned}
\mathcal{L}_o = \mathcal{L}_{\text{match}} + \gamma_{o} \cdot \mathcal{L}_{\text{overlap}},
\end{aligned}
\end{equation}
where 
\begin{equation}
    \begin{aligned}
        \mathcal{L}_{\text{match}} = -\frac{1}{N} \sum_{i=1}^{N} y_i^{gt} \log (y_i^{pred}),       
    \end{aligned}
\end{equation}
this improves class prediction accuracy for joint feature distributions, and
\begin{equation}
    \begin{aligned}
       \mathcal{L}_{\text{overlap}} = \frac{1}{\beta_{o} N} \sum_{i=1}^N \sum_{l \neq k} \log\left(e^{\beta_{o} y_{i,l}^{\text{pred}}} +e^{-\beta_{o} y_{i,k}^{\text{pred}}}\right),
    \end{aligned}
\end{equation}
in which $y_{i,k}^{\text{pred}}$ represents the prediction probability for the ground truth label $k$ of the $i$-th sample and $y_{i,l}^{\text{pred}}$ denotes the predicted probability for the other classes of the $i$-th sample. This loss penalizes inter-class feature overlap. The temperature parameter  $\beta_{o}$ controls overlap sensitivity, with higher values increasing inter-class margin enforcement. The hyper-parameter $\gamma_{o}$ balances label matching precision and class overlap suppression.

In this work, we refine the pseudo-label training with original adaptation loss and overlap loss during the UDA training, which helps minimize feature overlap for different classes. As illustrated in Fig.~\ref{Fig: OverlapLoss}, the attention-guided pseudo-labeling module is used to predict class distributions and pseudo-labels of the mapping features $f_{\mathcal{T} \rightarrow \mathcal{S}}$, which are complementary to source domain features. In contrast, the labels of mapping features $f_{\mathcal{S} \rightarrow \mathcal{T}}$ remain identical to source features $f_\mathcal{S}$  to guide overlap learning across domains.

\begin{figure*}[t]
    \centering
    \includegraphics[width=0.75\linewidth]{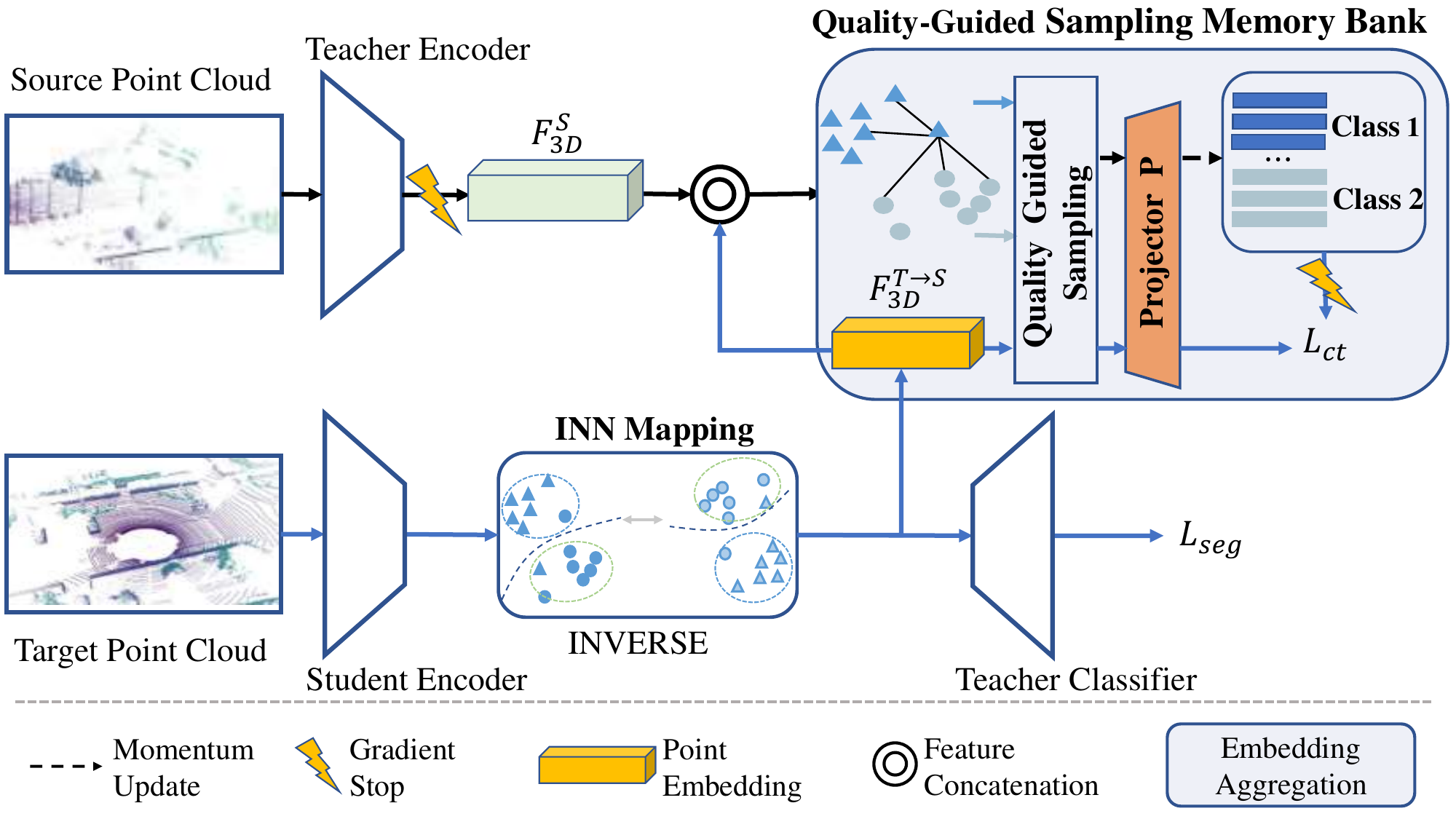}
    \caption{Illustration of our Quality-Guided Sampling Memory Bank, which selects high-quality and clean features for contrastive learning. By leveraging Quality-Guided Feature Sampling, the model effectively identifies and prioritizes discriminative features, reducing the impact of noisy pseudo-labels.}
    \label{Fig: Contrastive Memory Bank}
    \vspace{-3mm}
\end{figure*}

\subsection{Quality-Aware Contrastive Memory Bank}
While IAAM effectively ensures discriminative feature mapping and reduces feature overlap across domains, it does not explicitly address the issue of pseudo-label reliability, which remains a critical factor for transfer robustness. Building upon discriminative feature learning principles~\cite{xiao20233d,bashivan2021adversarial}, to further mitigate noisy pseudo-labels and cross-domain discrepancies, we propose a quality-aware contrastive memory bank comprising of: (1) adaptive quality-guided sampling for noise-robust feature selection, and (2) momentum-enhanced prototype contrastive learning for discriminative adaptation.

\noindent \textbf{Quality-guided sampling.} We utilize the quality-guided sampling module to dynamically filter out noise and prioritize high-quality features in the latent space. To compute the quality scores, we take into account both the density and cleanliness scores, which can be denoted as:
\begin{equation}
    \begin{aligned}
        \mathbf{q}(\mathbf{x}_i) 
        &= \frac{1}{k} \sum_{\mathbf{x}_j \in \mathcal{N}_k(\mathbf{x}_i)} \frac{1}{\|\mathbf{x}_i - \mathbf{x}_j\|_2+1} \\
        &- \frac{\gamma_{q}}{k} \sum_{\mathbf{x}_j \in \mathcal{N}_k(\mathbf{x}_i), \, y_j \neq y_i} \frac{\mathbf{x}_i \cdot \mathbf{x}_j}{\|\mathbf{x}_i\|_2 \|\mathbf{x}_j\|_2},
    \end{aligned}
\end{equation}
where $\mathcal{N}_k(\cdot)$ retrieves k nearest neighbors in the latent space and $\gamma_{q}$ balances density and cleanliness scoring which is empirically set to 1.0 in our experiment. Specifically, the density score quantifies the local concentration of similar features to promote robust clusters, while the cleanliness score penalizes the presence of neighboring features from different classes to ensure feature purity. For imbalance class weight input, we adaptively select neighbors by class-aware scaling:
\begin{equation}
k_c = k_{\text{base}} \cdot \sqrt[3]{\frac{N_{c}}{N_\text{total}}},
\end{equation}
where $N_c$ is the class population, the $k$ adaptively increases for head classes while reducing for tail classes, mitigating class imbalance in different scenarios. 

\noindent \textbf{Momentum contrastive learning.}
To further improve the quality of pseudo-label generation and improve the discriminability of class-wise features, we introduce a contrastive learning memory bank. This memory bank ensures that the model learns more structural and high-confidence representations by storing high-quality class-specific anchor features. Similar to~\cite{xiao20233d}, selected features undergo a projection module $\mathcal{P}: \mathbb{R}^d\rightarrow\mathbb{R}^m$ before memory integration. The memory bank $\mathcal{M}\in \mathbb{R}^{C \times M \times m}$ maintains $M$ prototypes per class through momentum updating, with updating rate $\beta_{m}$ and class-wise updating embeddings $\bar{\mathbf{z}}_c^t$:
\begin{equation}
\mathcal{M}_c^{t+1} = \beta_{m} \cdot\mathcal{M}_c^t + (1-\beta_{m}) \cdot \mathcal{P}(\bar{\mathbf{z}}_c^t).
\end{equation}

The contrastive learning objective enforces structural feature discrimination through:
\begin{equation}
    \begin{aligned}
        \mathcal{L}_{\text{con}} = -\frac{1}{N}\sum_{i=1}^N \log\left[ \frac{\sum_{p=1}^P e^{s_{ip}^+/\tau}}{\sum_{p=1}^P e^{s_{ip}^+/\tau} + \sum_{q=1}^Q e^{s_{iq}^-/\tau}} \right],
    \end{aligned}
\end{equation}
in which $s_{ip}^{+} = \text{sim}(\mathbf{z}_i, \mathbf{m}_{p}^{+})$ denotes the cosine similarity of the p-th positive prototype. In contrast, $s_{iq}^{-} = \text{sim}(\mathbf{z}_i, \mathbf{m}_{q}^{-})$ denotes the cosine similarity of the q-th negative prototype. The temperature parameter $\tau$ scales the logits before the softmax operation, controlling the sharpness of the similarity distribution. Empirically, we set $\tau=0.07$ to balance feature separation and stability in contrastive learning.

As shown in Fig. \ref{Fig: Contrastive Memory Bank}, the contrastive learning module is used to pull the embedding inputs of the same classes closer to the stored anchor embeddings while pushing away from others, encouraging learning from high-quality and more discriminative features. In the update stage, the memory bank is gradually updated with the high-quality fused features  $f_\mathcal{S}$ and $f_{\mathcal{T} \rightarrow \mathcal{S}}$, note that the student classifier is used to predict pseudo-labels for the mapping features and guide high-quality sampling during the update stage of the memory bank. In the contrastive learning stage, the mapping feature $f_{\mathcal{T} \rightarrow \mathcal{S}}$ is guided to align with high-quality stored anchor features and combined with the supervised segmentation loss $\mathcal{L}_{seg}$ during adaptation. The overall objective function for the memory bank training is:
\begin{equation}
    \begin{aligned}
        \mathcal{L}_c = \mathcal{L}_\text{seg} + \lambda_c \cdot \mathcal{L}_\text{con},
    \end{aligned}
\end{equation}
where the hyper-parameter $\lambda_c$ balances the contrastive learning and semantic segmentation learning.

\renewcommand{\arraystretch}{1.3}
\begin{table*}[ht]
\center
\caption{Quantitative class-wise comparisons for $\text{SynLiDAR} \rightarrow \text{SemanticPOSS}$. Target samples are adversarially perturbed with $\alpha = \text{0.03, 0.05, 0.07, 0.10}$ under I-FGSM attack and PGD attack for 10 iterations. The results are presented in terms of mean Intersection over Union (mIoU) and reported as percentages (\%).}
\resizebox{1.\linewidth}{!}{\noindent
\begin{tabular}{c|c|c|ccccccccccccc>{\columncolor{gray!30}}c}
\hline
\hline
  \multirow{1}{*}{\textbf{Adv. Obj.}} &
  \multicolumn{1}{c|}{\centering\parbox[t]{2mm}{\rotatebox{90}{attack}}} & 
  \multicolumn{1}{c|}{\parbox[t]{2mm}{$\alpha$}} & 
  \multicolumn{1}{c|}{\parbox[t]{2mm}{\rotatebox{90}{groun.}}} & 
  \multicolumn{1}{c|}{\parbox[t]{2mm}{\rotatebox{90}{buil.}}} & 
  \multicolumn{1}{c|}{\parbox[t]{2mm}{\rotatebox{90}{plants}}} &
  \multicolumn{1}{c|}{\parbox[t]{2mm}{\rotatebox{90}{fence}}} & 
  \multicolumn{1}{c|}{\parbox[t]{2mm}{\rotatebox{90}{rider}}} & 
  \multicolumn{1}{c|}{\parbox[t]{2mm}{\rotatebox{90}{car}}} & 
  \multicolumn{1}{c|}{\parbox[t]{2mm}{\rotatebox{90}{pole}}} & 
  \multicolumn{1}{c|}{\parbox[t]{2mm}{\rotatebox{90}{trunk}}} & 
  \multicolumn{1}{c|}{\parbox[t]{2mm}{\rotatebox{90}{pers.}}} & 
  \multicolumn{1}{c|}{\parbox[t]{2mm}{\rotatebox{90}{garb.}}} & 
  \multicolumn{1}{c|}{\parbox[t]{2mm}{\rotatebox{90}{traf.}}} & 
  \multicolumn{1}{c|}{\parbox[t]{2mm}{\rotatebox{90}{bike}}} & 
  \multicolumn{1}{c|}{\parbox[t]{2mm}{\rotatebox{90}{cone.}}} & 
  \multicolumn{1}{c|}{\rotatebox{0}{mIoU}}
  \Tstrut\Bstrut\\

\hline
\multirow{1}{*}{CosMix~\cite{saltori2023compositional}} 
&-   &-  &82.62	&66.47	&69.09	&32.09	&51.39	&38.19	&36.52	&22.54	&56.44	&27.25	&19.84	&5.77	&26.31 &41.12\\
\hline
\multirow{4}{*}{PGD~\cite{xu2023adversarial}} 
&\cmark &0.03 &81.94 &63.16 &68.04 &33.30 &50.00 &38.34 &37.14 &21.87 &54.93 &27.86 &18.69 &5.09 &24.67 &40.39\\
&\cmark &0.05 &81.44 &56.82	&65.67 &34.28 &48.13 &38.63	&37.79 &20.82 &52.92 &24.81	&17.55 &4.57 &24.00 &39.03\\
&\cmark &0.07 &80.43 &49.82	&62.83 &33.40 &45.67 &36.32	&36.59 &19.43 &50.34 &18.01	&15.74 &4.06 &23.08 &36.59\\
&\cmark &0.10 &76.83 &42.75 &58.85 &24.51 &39.76 &30.33 &31.57 &18.15 &46.33 &9.59 &12.98 &3.32 &20.15 &31.93\\
\hline
\multirow{4}{*}{I-FGSM~\cite{liu2019extending}} 
&\cmark &0.03   &81.93	&63.13	&68.05	&33.24	&49.98	&38.37	&37.23	&21.86	&54.94	&28.12	&18.55	&5.09	&24.46   &40.38\\
&\cmark   &0.05   &81.44	&56.80	&65.68	&34.26	&47.88	&38.51	&37.60	&20.78	&52.88	&24.95	&17.55	&4.59	&24.05   &39.00\\
&\cmark   &0.07   &80.42	&49.83	&62.82	&33.44	&45.52	&36.33	&36.50	&19.38	&50.24	&17.93	&15.59	&4.06	&23.04   &36.55\\
&\cmark   &0.1    &76.83	&42.76	&58.87	&24.51	&39.67	&30.41	&31.55	&18.10	&46.23	&9.60	&12.96	&3.31	&20.06    &31.91\\
\hline
\hline
\end{tabular}
}
\label{table:4-1 lidar2poss under attack}
\end{table*}

\renewcommand{\arraystretch}{1.3}
\begin{table*}[ht]
\center
\caption{Quantitative class-wise comparisons when using IAAM and QC-MB for $\text{SynLiDAR} \rightarrow \text{SemanticPOSS}$, we conduct robustness evaluation under PGD and I-FGSM attack with perturbation intensity $\alpha=10$. The robustness scores (rb.dr) are reported as percentages (\%). Bold fonts indicate the highest performance under attack.}
\resizebox{1.\linewidth}{!}{\noindent
\begin{tabular}{c|c|c|c|cccccccccccccc>{\columncolor{gray!30}}c}
\hline
\hline
  \multirow{1}{*}{\textbf{Adv. Obj.}} &
  \multicolumn{1}{c|}{\centering\parbox[t]{2mm}{\rotatebox{90}{attack}}} &
  \multicolumn{1}{c|}{\centering\parbox[t]{2mm}{\rotatebox{90}{IAAM}}} &
  \multicolumn{1}{c|}{\centering\parbox[t]{2mm}{\rotatebox{90}{QC-MB}}} & 
  \multicolumn{1}{c|}{\parbox[t]{2mm}{\rotatebox{90}{groun.}}} & 
  \multicolumn{1}{c|}{\parbox[t]{2mm}{\rotatebox{90}{buil.}}} & 
  \multicolumn{1}{c|}{\parbox[t]{2mm}{\rotatebox{90}{plants}}} &
  \multicolumn{1}{c|}{\parbox[t]{2mm}{\rotatebox{90}{fence}}} & 
  \multicolumn{1}{c|}{\parbox[t]{2mm}{\rotatebox{90}{rider}}} & 
  \multicolumn{1}{c|}{\parbox[t]{2mm}{\rotatebox{90}{car}}} & 
  \multicolumn{1}{c|}{\parbox[t]{2mm}{\rotatebox{90}{pole}}} & 
  \multicolumn{1}{c|}{\parbox[t]{2mm}{\rotatebox{90}{trunk}}} & 
  \multicolumn{1}{c|}{\parbox[t]{2mm}{\rotatebox{90}{pers.}}} & 
  \multicolumn{1}{c|}{\parbox[t]{2mm}{\rotatebox{90}{garb.}}} & 
  \multicolumn{1}{c|}{\parbox[t]{2mm}{\rotatebox{90}{traf.}}} & 
  \multicolumn{1}{c|}{\parbox[t]{2mm}{\rotatebox{90}{bike}}} & 
  \multicolumn{1}{c|}{\parbox[t]{2mm}{\rotatebox{90}{cone.}}} & 
  \multicolumn{1}{c|}{\rotatebox{0}{mIoU}} &
  \multicolumn{1}{c|}{\rotatebox{0}{rb.dr\%}}
\Tstrut\Bstrut\\

\hline
\multirow{1}{*}{CosMix~\cite{saltori2023compositional}} 
&-   &- & - &82.62	&66.47	&69.09	&32.09	&51.39	&38.19	&36.52	&22.54	&56.44	&27.25	&19.84	&5.77	&26.31 &41.12 & -\\
\hline
\multirow{3}{*}{PGD~\cite{xu2023adversarial}} 
&\cmark & - & - &76.83	&42.75	&58.85	&24.51	&39.76	&30.33	&31.57	&18.15	&46.33	&9.59	&12.98	&3.32	&20.15 &31.93 & 22.35\\
&\cmark &\cmark & - &80.64	&47.78	&63.16	&25.40	&42.29	&34.46	&39.58	&20.97	&46.61	&12.10	&17.99	&7.81	&17.75 &35.12 & 14.59\\
&\cmark &\cmark &\cmark &80.25	&65.75	&69.53	&24.66	&45.84	&37.37	&33.93	&21.41	&48.43	&16.31	&22.37	&7.46	&18.23 &\bf{37.81} &8.05\\
\hline
\multirow{3}{*}{I-FGSM~\cite{liu2019extending}} 
&\cmark & - & - &76.83	&42.76	&58.87	&24.51	&39.67	&30.41	&31.55	&18.10	&46.23	&9.60	&12.96	&3.31	&20.06 &31.91 &22.39\\
&\cmark &\cmark & - &80.63	&47.76	&63.15	&25.43	&42.27	&34.51	&39.38	&21.05	&46.68	&12.00	&18.13	&7.80	&17.60 &35.11 &14.62 \\
&\cmark &\cmark &\cmark &80.25	&65.75	&69.53	&24.65	&45.74	&37.37	&33.95	&21.31	&48.35	&15.82	&22.11	&7.46	&18.32 &\bf{37.74} &8.22\\

\hline
\hline
\end{tabular}
}
\label{table:4-2 lidar2poss robustness}
\end{table*}

\section{Experiment}
In this section, we present a comprehensive evaluation of unsupervised domain adaptation on 3D point cloud semantic segmentation (PCSS-UDA) robustness. Section $\text{IV-A}$ details the model implementation and experiment settings. Our study compares synthetic-to-real adaptation using the SynLiDAR~\cite{xiao2107learning} and SemanticPOSS~\cite{pan2020semanticposs} datasets, and real-to-real domain generalization using 
SemanticKITTI~\cite{behley2019semantickitti} and SemanticSTF~\cite{xiao20233d} under adversarial weather conditions. Furthermore, we analyze the impact of the proposed robustness components through an ablation study, demonstrating their independent contributions to improve the robustness of PCSS-UDA.

\subsection{Implementation Details}
We use SynLiDAR as the synthetic dataset for synthetic-to-real adaptation and SemanticPOSS as the real-world dataset. SynLiDAR is a large-scale synthetic dataset comprising 198,396 annotated point clouds with 32 semantic classes, while SemanticPOSS consists of 2,988 annotated real-world point clouds and 14 semantic labels. For real-to-real domain adaptation, we employ SemanticKITTI, a widely used large-scale dataset that provides point-wise semantic annotations of the KITTI dataset~\cite{geiger2013vision}. In addition, we use the SemanticSTF dataset, which contains 2,076 manually annotated scans captured by the Velodyne HiDL64S3D LiDAR sensor from STF~\cite{bijelic2020seeing} dataset. SemanticSTF includes various adverse weather conditions, such as snow, dense fog, light fog, and rain. Following CosMix \cite{saltori2023compositional}, we adopt the same training and validation split for $ \text{SynLiDAR} \rightarrow \text{SemanticPOSS}$, while for $\text{SemanticKITTI} \rightarrow \text{SemanticSTF}$, we use the original validation split. We employ the MinkUnet34 3D sparse convolution model~\cite{choy20194d} as our backbone architecture. The hyper-parameter $\gamma_{o}$ and $\beta_{o}$ are empirically set to 1.0 and 20.0 for overlap suppression. During contrastive learning, we empirically set the number of class anchors stored in the contrastive memory bank to 64 to optimize memory usage and computational efficiency and set $\beta_m$ to 0.98 for memory update. All experiments are conducted using the PyTorch~\cite{paszke2019pytorch} framework on 2 $\times$ NVIDIA GeForce RTX4090. 

\begin{figure}[t]
    \centering
    \includegraphics[width=0.99\linewidth]{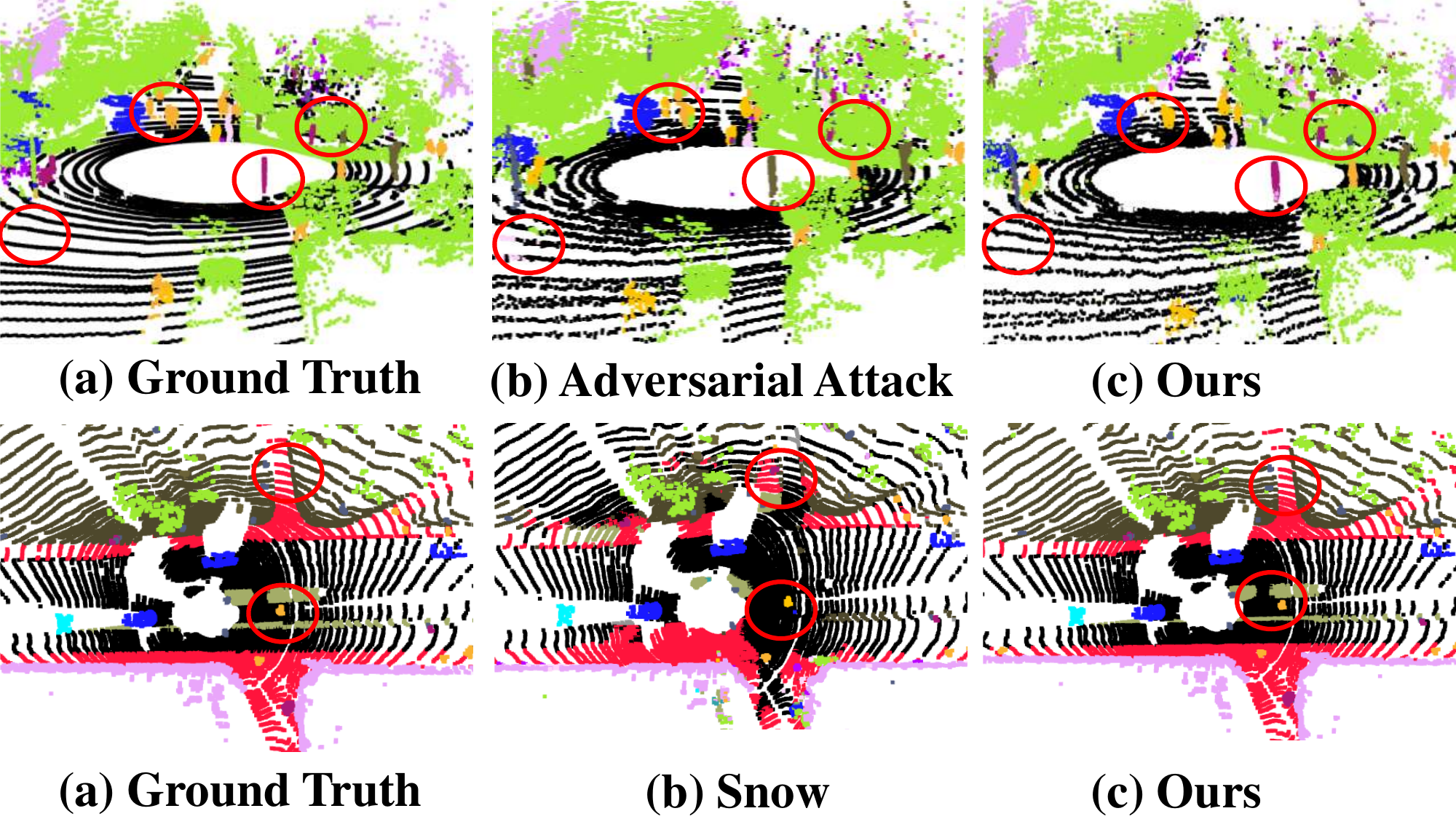}
    \caption{Illustration of PCSS-UDA with our methods under adversarial conditions. In the top row, (a) shows the ground truth, (b) adversarial attack degradation, and (c) highlights our method’s improved performance by prioritizing high-quality features. In the bottom row, (a) shows the ground truth under snow conditions, (b) illustrates domain shift challenges, and (c) shows how our methods work.}
    \label{Fig: Experiments}
    \vspace{-3mm}
\end{figure}

\subsection{Performance and Comparisons}
\noindent \textbf{Adversarial robustness.} To systematically evaluate adversarial robustness, we benchmark models trained on $\text{SynLiDAR}\rightarrow\text{SemanticPOSS}$ under I-FGSM and PGD attack with increasing perturbation intensities ($\alpha = \text{0.03, 0.05, 0.07, 0.10}$). The comparative results in Table.~\ref{table:4-1 lidar2poss under attack} present the class-wise IoU performance. Both types of attacks exhibit a strong positive correlation between $\alpha$ and performance drop. At $\alpha = 0.10$, the PGD attack reduces mIoU to 31.93, which is a 22.35\% decrease from clean target data, demonstrating that the PCSS-UDA model is severely vulnerable to high-intensity perturbations. Structural classes (buildings/plants) show a disproportionate sensitivity, with the IoU of buildings dropping from 66.47 (clean) to 42.75 under PGD-0.10 ($\Delta = 23.72$), probably because point injections disrupt their geometric regularity. In contrast, small movable objects (bikes) suffer near complete failure ($\leq$ 5 IoU), indicating the catastrophic collapse of the feature space.

\noindent \textbf{Component effectiveness.} Table.~\ref{table:4-2 lidar2poss robustness} quantitatively dissects the adversarial robustness gains from our proposed modules. The integration of IAAM achieves 35.12 mIoU under PGD-$\alpha = 0.1$ attack, with an improvement in absolute robustness of 7.75\% over the baseline and leads to improvement in performance across multiple classes. This suggests that the improvement in robustness can be attributed to the model's ability to handle feature overlap during adaptation and dynamically adjust decision boundaries across various classes. For example, classes with complex structure such as plants, buildings, and tail classes (e.g., rider, car, pole) show substantial improvement under PGD and I-FGSM attacks. Integrating the framework with the QC-MB further improves mIoU to 37.81, particularly for building, rider, car, and other tail classes. Performance improvements in the building are significant, with the IoU increasing from 42.75 to 65.75 under PGD attack. Tail classes (e.g., rider, car, traffic signs) also show relative improvement. 
As shown in Fig.~\ref{Fig: Experiments}, with adversarial point cloud perturbations, our model can distinguish structural classes and reduce the overlap (e.g., the mixture of ground and other classes). In the top row, the model still manages to separate features from different classes, particularly in the structural classes under adversarial attack. Similarly, under domain shift conditions (snow) in the bottom row, the model successfully minimizes the class overlap and maintains class-specific feature boundaries. This suggests that the QC-MB helps distill discriminative features during adaptation.

\renewcommand{\arraystretch}{1.3}
\begin{table}[t]
\center
\caption{Ablation experiments with the components of IAAM and QC-MB, in which $\text{L} \rightarrow \text{P}$ represents PCSS-UDA for $\text{SynLiDAR}\rightarrow\text{SemanticPOSS}$ and  $\text{K} \rightarrow \text{STF}$ represents PCSS-UDA for $\text{SemanticKITTI}\rightarrow\text{SemanticSTF}$. The results are presented in terms of mean Intersection over Union (mIoU) and reported as percentages (\%). }
\resizebox{0.8\linewidth}{!}{\noindent
\begin{tabular}{c|c|c|c|c|c}
\hline
\hline
  \multirow{2}{*}{Methods} & 
  \multirow{2}{*}{attack} &
  \multirow{2}{*}{IAAM} &
  \multirow{2}{*}{QC-MB} &
  \multicolumn{2}{c}{mIoU} \Tstrut\Bstrut\\
  \cline{5-6}
  & & & & $\text{L} \rightarrow \text{P}$ & $\text{K} \rightarrow \text{STF}$
  \Tstrut\Bstrut\\
\hline
CosMix
& - & - & - & 41.12 & 30.12\Tstrut\\
\hline
(a)
&  & \cmark & & 40.70 & -\Tstrut\\
(b)
&  & \cmark & \cmark & 40.91 & 30.37\Tstrut\\
(c)
& \cmark &  &  & 31.93 & - \Tstrut\\
(d)
& \cmark & \cmark & & 35.12 & -\Tstrut\\
(e)
& \cmark & \cmark &  \cmark & 37.91 & -\Tstrut\\

\hline
\hline
\end{tabular}}
\label{table:4-ablation}
\vspace{-5mm}
\end{table}

\begin{figure}[t]
    \centering
    \includegraphics[width=0.95\linewidth]{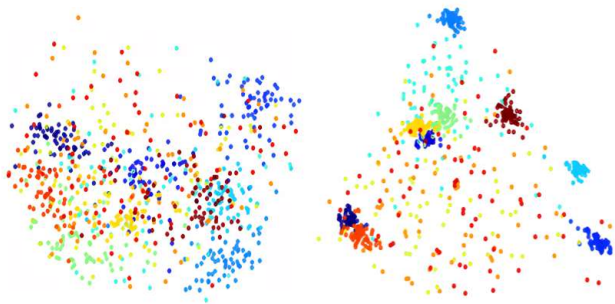}
    \caption{Illustration of the Isomap feature space representations in the Contrastive Learning Memory Bank. The left side displays the feature distribution before contrastive learning, showing more overlap across classes. On the right, post-contrastive learning, the features are better separated, with improved cluster alignment for different classes. }
    \label{Fig: Contrastive Memory Bank Features}
    \vspace{-5mm}
\end{figure}

\subsection{Ablation Studies}
The ablation study shown in Table.~\ref{table:4-ablation} further explores the effects of IAAM and QC-MB. Introducing IAAM results in a slight drop in clean performance (40.70 mIoU in experiment (a)) but significantly improves adversarial robustness (35.12 mIoU in experiment (d)). Combining IAAM with QC-MB (experiment (b)) shows improvements in clean and adversarial conditions, reaching 37.91 mIoU. In domain generalization tasks, the proposed methods improve the mIoU to 30.37 for $\text{SemanticKITTI} \rightarrow \text{SemanticSTF}$, which is 0.66\% higher than the CosMix base. Feature visualizations in Fig.~\ref{Fig: Contrastive Memory Bank Features} highlight the improved cluster separation after contrastive learning, reducing the overlap in classes and improving domain generalization. However, for points with very low frequency in the training dataset, the degree of feature space separation is still insufficient, which may be related to our down-sampling approach.

\section{Conclusions}
In this paper, we propose a novel framework to improve the robustness of PCSS-UDA models against adversarial attacks and common corruption. First, we propose an INN-based invertible mapping model for bidirectional alignment and reduce feature overlap across domains with overlap suppression loss. In addition, we introduce a quality-guided contrastive memory bank (QC-MB) to distill high-quality target features during adaptation, thus exploiting more discriminative target features and progressively refining pseudo-label generalization. Our extensive experimental results demonstrate that the proposed methods significantly improve performance under I-FGSM and PGD attacks, as well as domain generalization under adversarial weather conditions. These results suggest that minimizing feature overlap and preserving discriminative feature structure are crucial for improving the generalization and robustness of PCSS-UDA models. To this end, our framework provides a comprehensive benchmark for evaluating robustness in PCSS-UDA and offers a promising direction for future research in adversarial robustness of PCSS-UDA tasks.

{\small
\bibliographystyle{IEEEtran}
\bibliography{root}
}
\end{document}